\documentclass{article}

\usepackage[preprint,nonatbib]{neurips_2019}

\usepackage[utf8]{inputenc} 
\usepackage[T1]{fontenc}    
\usepackage{hyperref}       
\usepackage{url}            
\usepackage{booktabs}       
\usepackage{amsfonts}       
\usepackage{nicefrac}       
\usepackage{microtype}      
\usepackage{amsmath}
\title{Sample Complexity of an Adversarial Attack on UCB-based Best-arm Identification Policy }

\begin{document}
\author{Varsha Pendyala\\ Dept. of Electrical and Computer Engineering\\ University of Wisconsin-Madison\\ pendyala@wisc.edu}
\maketitle
\begin{abstract}
  In this work I study the problem of adversarial perturbations to rewards, in a Multi-armed bandit (MAB) setting. Specifically, I focus on an adversarial attack to a UCB type best-arm identification policy applied to a stochastic MAB. The UCB attack presented in [1] results in pulling a target arm K very often. I used the attack model of [1] to derive the sample complexity required for selecting target arm K as the best arm. I have proved that the stopping condition of UCB based best-arm identification algorithm given in [2], can be achieved by the target arm K in T rounds, where T depends only on the total number of arms and $\sigma$ parameter of $\sigma^2-$ sub-Gaussian random rewards of the arms.
\end{abstract}
\section{Introduction}
Multi-armed bandits (MAB) are a class of problems where the learner is tasked with allocating available resources among various alternatives (arms) in an online setting with limited feedback. At each time step, learner allocates a unit resource and receives a payoff based on the chosen arm. By the nature of the reward generation process, mainly there are two different formalizations of the bandit problem - stochastic, adversarial [3]. In a stochastic MAB, rewards are generated independently at random from underlying fixed unknown probability distributions with means $\mu_i$ for each arm i=1,2,…,K. In an adversarial bandit problem, rewards may be selected arbitrarily by an adversary. 

In addition, depending upon the goal of learner, there are several sub-classes of problems. Regret minimizing bandit problems focus on devising schemes which result in gains comparable to that of best policy, which pulls single best arm in hindsight always. Thus, they focus on reducing the notion of regret $R_n = n\mu_* - E(\sum_{t=1}^{n}X_t)$ ,where $\mu_*$  is the mean of the arm with highest expected reward. Another important class is best-arm identification problem, where the goal of bandit learner is to identify the arm with highest mean in a stochastic MAB problem. 

Stochastic MABs are widely used in important applications in various industry segments[6] [7]– operations research, ad-recommendations, medical treatment planning, among many others. Stochastic MAB like many other machine learning algorithms can be vulnerable to adversarial attacks. In this project, I focus on an adversarial attack to a best-arm identification problem using the approach described in [1] for generating adversarial perturbations. I derive the sample complexity for choosing the target arm as the best arm when the learner uses UCB [2][3] type algorithm for best-arm identification.

In [1], authors describe an adversarial attack to UCB bandit learner that results in pulling a target arm K very often. It is achieved by introducing perturbations $\alpha_t$ to random rewards $r_t^0$ generated from underlying distribution, thereby manipulating the reward to $r_t = r_t^0 - \alpha_t$. In section 2, I introduce the notation from [1] and briefly describe the corresponding attack process. I then elaborate on the best-arm identification policy of [2] in section 3. Derivation of the sample complexity for target arm selection is presented in section 4. Sections 5 and 6 has the description of related work and conclusions respectively. 
\section{Attack on UCB learner}
It is assumed that all the arm rewards are $\sigma^2-$sub-Gaussian, where $\sigma^2$ is known to both attacker and learner.\\
\\
\textbf{Notation:}
\begin{align*}
    \tau_i(t) &:= \{s:s\leq t, I_s = i\}, \text{ set of rounds up to $t$ where arm $i$ is chosen}\\
    \hat{\mu}^0_i(t) &:= N_i(t)^{-1}\sum_{s\in \tau_i(t)}r_s^0, \text{ pre-attack empirical mean of the rewards of arm $i$ up to time $t$}\\
    \hat{\mu}_i(t) &:= N_i(t)^{-1}\sum_{s\in \tau_i(t)}r_s, \text{ post-attack empirical mean of the rewards of arm $i$ up to time $t$}\\
    \beta(N)&:=\sqrt{\frac{2\sigma^2}{N}\log{\frac{\pi^2 K N^2}{3\delta}}}\\
    \Delta_i &:= max\{\mu_i - \mu_K,0\}
\end{align*}

Arm selection rule of UCB [1]:
\[
I_t = 
\begin{cases}
t & \text{if } t\leq K\\
argmax_{i}\left\{\hat{\mu}_i(t-1)+3\sigma \sqrt{\frac{\log{t}}{N_i(t-1)}}\right\} & \text{otherwise}
\end{cases}
\]
Learner plays each arm once in the first K rounds during which no attack is made $(\alpha_t=0)$. In the subsequent rounds, attack is made only when $I_t=i\neq K$. It follows from the arm selection rule that
$$\hat{\mu}_i(t-1)+3\sigma \sqrt{\frac{\log{t}}{N_i(t-1)}} \geq \hat{\mu}_K(t-1)+3\sigma \sqrt{\frac{\log{t}}{N_K(t-1)}}$$
An attack $\alpha_t$  with smallest absolute value is computed such that
$$\hat{\mu}_i(t)\leq \hat{\mu}_K(t-1)-2\beta (N_K(t-1))-\Delta_0$$
where $\Delta_0 \geq 0$ is a parameter of the attacker. Empirical mean post-attack is computed as follows 
$$\hat{\mu}_i(t) = \frac{N_i(t-1)\hat{\mu}_i(t-1)+r_t^0-\alpha_t}{N_i(t-1)+1}$$
closed form attack $\alpha_t$:
$$\alpha_t = \left[N_i(t)\hat{\mu}_i^0(t)-\sum_{s\in \tau_i(t-1)}\alpha_s-N_i(t).(\hat{\mu}_K(t-1)-2\beta(N_K(t-1))-\Delta_0)\right]_+$$\\
Following theorem is proved in [1]:\\
\\
\textbf{Theorem 1:} Suppose $T\geq 2K$ and $\delta \leq 1/2$. The with probability at least $1-\delta$, attacker forces bandit learner to choose the target arm in at least 
$$T-(K-1)\left(2+\frac{9\sigma^2}{\Delta_0^2}\log{T}\right)$$
rounds, using a cumulative attack cost at most
$$\sum_{t=1}^{T}\alpha_t \leq \left(2+\frac{9\sigma^2}{\Delta_0^2}\log{T}\right) \sum_{i<K}(\Delta_i + \Delta_0)+\sigma (K-1)\sqrt{32(2+\frac{9\sigma^2}{\Delta_0^2}\log{T})\log{\frac{\pi^2 K (2+\frac{9\sigma^2}{\Delta_0^2}\log{T})^2}{3\delta}}}$$
This comes from the following lemma, also proved in [1]:\\
\\
\textbf{Lemma 1:} : When event E holds and $\delta \leq 1/2$. Then for any $i<K$ and $t\geq 2K$, we have 
$$N_i(t)\leq min\left\{N_K(t),2+\frac{9\sigma^2}{\Delta_0^2}\log{T}\right\}$$
where $E:=\left\{\forall i, \forall t >K: |\hat{\mu}^0_i(t)-\mu_i|<\beta (N_i(t)) \right\}$
\section{Best-arm identification using UCB}
The algorithm samples the arm indexed by 
$$argmax_{i\in[n]}\hat{\mu}_{i,T_i(t)}+C_{i,t}$$
where $C_{i,t}>0$ is typically derived from a tail bound.
Stopping condition for choosing the best arm is given by [2][4] :
\begin{align}
    \exists i \in [n]: T_i(t) \geq \alpha \sum_{j\neq i}T_j(t)
\end{align}
and output $argmax_i T_i (t)$ for some $\alpha > 0$

\subsection{Validating the above stopping condition:}
Following a similar analysis as [2], I validate the stopping condition (1) for this UCB problem:\\
Using the Markov’s inequality and $\psi(\epsilon) = \frac{\epsilon^2}{2\sigma^2}$, from [3] we have with probability at least $1-\delta$:
$$|\hat{\mu}_{i,T_i(t)} - \mu_i|\leq (\psi)^{-1}(\frac{1}{T_i(t)}\ln{\frac{1}{\delta}})$$
Let $\delta = t^{-\alpha}$. It implies, with probability at least $1-\delta$
$$|\hat{\mu}_{i,T_i(t)} - \mu_i|\leq \sigma \sqrt{\frac{2\alpha\ln{t}}{T_i(t)}}$$
Assume $i_*$ is the best arm and $C_{i,t}=(1+\beta)\left(\sigma \sqrt{\frac{2\alpha\ln{t}}{T_i(t)}}\right)$ for some $\beta >0$.
If UCB chooses sub-optimal arm $i$, then following inequalities hold with at least $1-\delta$ probability:
\begin{align*}
    \mu_i + (2+\beta)\sigma \sqrt{\frac{2\alpha\ln{t}}{T_i(t)}} &\geq \hat{\mu}_{i,T_i(t)} + (1+\beta) \sigma \sqrt{\frac{2\alpha\ln{t}}{T_i(t)}}\\
    &\geq \hat{\mu}_{i_*,T_{i_*}(t)} + (1+\beta) \sigma \sqrt{\frac{2\alpha\ln{t}}{T_{i_*}(t)}} \\
    &\geq \mu_{i_*} + \beta \sigma \sqrt{\frac{2\alpha\ln{t}}{T_{i_*}(t)}}
\end{align*}
This implies, we have
$$(2+\beta)\sigma \sqrt{\frac{2\alpha\ln{t}}{T_i(t)}} \geq \beta \sigma \sqrt{\frac{2\alpha\ln{t}}{T_{i_*}(t)}}$$
On simplifying the above
\begin{align}
    T_i(t)\leq \left(\frac{2+\beta}{\beta}\right)^2 T_{i_*}(t)
\end{align}
Thus, any sub-optimal arm $i$ has $T_i(t) \leq \alpha T_{i_*}(t)$ with high probability, where $\alpha = \left(\frac{2+\beta}{\beta}\right)^2$

It implies, only $i_*$  can achieve the stopping condition (1) thus ensuring the choosing of optimal arm indeed by the UCB with high probability.
\section{Sample complexity for target arm selection}
In this section, I apply the adversarial attack described in section 2 to a best-arm identification setting. This is motivated by the fact that, when target arm K is pulled very often through reward manipulation, it will appear that the expected reward of arm K is highest among all. Hence, any valid best-arm identification algorithm should terminate with arm K as the best arm. 

From Theorem 1 we have a bound on the number of times a non-target arm $i$ is pulled up to time $t$.
\begin{align*}
    N_i(t) &\leq \left(2+\frac{9\sigma^2\log{t}}{\Delta^2}\right)\\
    \sum_{j\neq K}N_j(t) &\leq (K-1)\left(2+\frac{9\sigma^2\log{t}}{\Delta^2}\right)\\
    \alpha \sum_{j\neq K}N_j(t) &\leq \alpha (K-1)\left(2+\frac{9\sigma^2\log{t}}{\Delta^2}\right)\text{ ($\alpha > 0$)}\\
    N_{i_*}(t) &\geq t-(K-1)\left(2+\frac{9\sigma^2\log{t}}{\Delta^2}\right)
\end{align*}
\\
\\
To ensure that the arm K is chosen as the best arm using the stopping condition (1), following criteria has to be met:
\begin{align*}
    t-(K-1)\left(2+\frac{9\sigma^2\log{t}}{\Delta^2}\right) &\geq \alpha (K-1)\left(2+\frac{9\sigma^2\log{t}}{\Delta^2}\right)\\
    \frac{t}{\left(2+\frac{9\sigma^2\log{t}}{\Delta_0^2}\right )}  &\geq (\alpha + 1) (K-1)
\end{align*}
\begin{align}
    t - (\alpha +1)(K-1)\left(\frac{9\sigma^2\log{t}}{\Delta_0^2}\right) &\geq 2(\alpha + 1)(K-1) 
\end{align}

Let $f(t) =  t - (\alpha +1)(K-1)\left(\frac{9\sigma^2\log{t}}{\Delta_0^2}\right)$\\
Deriving the condition on $\Delta_0$ to make $f(t)$ a monotonically increasing function:
$$f^{'}(t) = 1 - (\alpha + 1)(K-1)\frac{9\sigma^2}{\Delta_0^2 t} $$
Clearly, $f^{'}(t)$ is an increasing function of $t.$ Hence, it is sufficient to ensure $f^{'}(1)$ is positive:
$$\Delta_0^2 > (\alpha +1)(K-1) 9 \sigma^2$$
From (2) we have $\alpha = \left(\frac{2+\beta}{\beta}\right)^2$
\begin{align}
   \implies \Delta_0 > 3\sigma \sqrt{(K-1)\left(1+\left(\frac{2+\beta}{\beta}\right)^2\right)}
\end{align}
By choosing $\Delta_0$ such that (4) is satisfied, UCB returns arm K as the best arm with high probability at round $t$ that satisfies (3).

From (3), we have that the sample complexity only depends upon $K, \sigma$ and $\beta$.

From Theorem 1, it implies the attack cost is [1] $$\hat{O}(\sum_{i<K}\Delta_i \log{t} + \sigma K \log{t})$$ 
where $\hat{O}$ ignores $\log{\log{t}}$ factors.
\section{Related work}
There is vast amount of literature on best-arm identification problem in a stochastic MAB setting. Several approaches cater to either of the two problem settings-one where the learner is given a fixed budget of arm pulls (\textit{fixed budget setting}) and the other where the learner has to attain a target confidence level in it's prediction (\textit{fixed confidence setting}) . Majority of the existing approaches like those presented in [2][5][8], result in a logarithmic multiplicative gap between the known lower and upper bound over the number of arm pulls. 

Towards the other end of the spectrum, several studies like [1][9][10] have explored on the adversarial attacks to machine learning algorithms which operate under stochastic assumptions. Recent work presented in [1] [11] brings the issue of adversarial attacks to MAB-like setting where the attacker perturbs reward signals to make the learner pull a target arm very often or visit a target state.

In this work, I have attempted to derive sample complexity of the attack to a best-arm identification policy using the attack strategy stated in [1].
\section{Conclusion}
I have derived an expression to compute the minimum rounds required for an adversary to mislead a UCB based best-arm identification learner into choosing the target arm as the best arm. I have proved that the sample complexity of this attack depends only on the total number of arms, $\sigma$ parameter of $\sigma^2-$sub-Gaussian random rewards and $\beta$ parameter of the UCB learner.
\section{References}
\small 

[1] Jun KS, Li L, Ma Y, Zhu J. Adversarial attacks on stochastic bandits. InAdvances in Neural Information Processing Systems 2018 (pp. 3640-3649).

[2] Jamieson, Kevin, and Robert Nowak. "Best-arm identification algorithms for multi-armed bandits in the fixed confidence setting." 2014 48th Annual Conference on Information Sciences and Systems (CISS). IEEE, 2014.

[3]  Bubeck, Sébastien, and Nicolo Cesa-Bianchi. "Regret analysis of stochastic and nonstochastic multi-armed bandit problems." Foundations and Trends® in Machine Learning 5.1 (2012): 1-122.

[4] Jamieson, Kevin, et al. "lil’ucb: An optimal exploration algorithm for multi-armed bandits." Conference on Learning Theory. 2014.

[5] Audibert, Jean-Yves, and Sébastien Bubeck. "Best arm identification in multi-armed bandits." COLT-23th Conference on learning theory-2010. 2010.

[6] Li, Lihong, Chu, Wei, Langford, John, and Schapire, Robert E. A contextual-bandit approach to personalized news article recommendation. In Proceedings of the Nineteenth International Conference on World Wide Web (WWW), pp. 661–670, 2010.

[7] Kuleshov, Volodymyr and Precup, Doina. Algorithms for multi-armed bandit problems. CoRR abs/1402.6028, 2014.

[8] Karnin, Zohar, Tomer Koren, and Oren Somekh. "Almost optimal exploration in multi-armed bandits." International Conference on Machine Learning. 2013.

[9] Pattanaik, Anay, et al. "Robust deep reinforcement learning with adversarial attacks." Proceedings of the 17th International Conference on Autonomous Agents and MultiAgent Systems. International Foundation for Autonomous Agents and Multiagent Systems, 2018.

[10] Goodfellow, Ian J, Shlens, Jonathon, and Szegedy, Christian. Explaining and harnessing adversarial
examples. In International Conference on Learning Representations, 2015.

[11] Lin, Yen-Chen, Hong, Zhang-Wei, Liao, Yuan-Hong, Shih, Meng-Li, Liu, Ming-Yu, and Sun,
Min. Tactics of adversarial attack on deep reinforcement learning agents. In Proceedings of the
26th International Joint Conference on Artificial Intelligence (IJCAI), pp. 3756–3762, 2017.

\end{document}